\documentclass{article}

\usepackage{acronym}
\acrodef{CCE}{colon capsule endoscopy}
\acrodef{CV}{cross-validation}
\acrodef{AI}{artificial intelligence}

\usepackage{mathtools}
\usepackage[a4paper, margin=2.45cm]{geometry}
\usepackage{graphicx}
\usepackage{siunitx}
\usepackage{booktabs}
\usepackage{bm}
\usepackage{pifont}
\usepackage[numbers]{natbib}
\usepackage{parskip}
\usepackage{titlesec}

\usepackage[T1]{fontenc}
\usepackage{lmodern}

\usepackage{hyperref}

\DeclarePairedDelimiter{\norm}{\lVert}{\rVert}

\title{Polyp Matching Using Deep Learning}

\author{%
  Kristian Dalsbø Hindberg\textsuperscript{1,\textasteriskcentered}
  \and
  Puneet Sharma\textsuperscript{2}
  \and
  Benedicte Schelde-Olesen\textsuperscript{3,4}
  \and
  Ulrik Deding\textsuperscript{3,4}
  \and
  Jakob Frederik Frøkjær Justsen\textsuperscript{3,4}
  \and
  Jan-Matthias Braun\textsuperscript{5,6}
  \and
  Esmaeil S. Nadimi\textsuperscript{5,6}
  \and
  All on behalf of the AICE Consortium\textsuperscript{1--6}
}

\date{}

\begin{document}
\maketitle

\noindent
\textsuperscript{1}Department of Physics and Technology, Machine Learning Group, UiT -- The Arctic University of Norway, Tromsø, Norway\\
\textsuperscript{2}Department of Automation and Process Engineering, Machine Learning Group, UiT -- The Arctic University of Norway, Tromsø, Norway\\
\textsuperscript{3}Department of Surgery, Odense University Hospital, Odense, Denmark\\
\textsuperscript{4}Department of Clinical Research, University of Southern Denmark, Odense, Denmark\\
\textsuperscript{5}Faculty of Engineering, University of Southern Denmark, Odense, Denmark\\
\textsuperscript{6}Applied AI and Data Science (AID), Maersk Mc-Kinney Moller Institute, Odense, Denmark

\noindent
\textsuperscript{\textasteriskcentered}Corresponding author: \texttt{kristian.d.hindberg@uit.no}

\begin{abstract}
\noindent A key challenge in \ac{CCE} analysis is accurately identifying whether multiple images depict the same polyp, which is critical to avoid over-counting and unnecessary clinical interventions. Additionally, this prevents prolonged follow-up colonoscopies where endoscopists search for duplicate polyps without precise localization. This study presents a novel contrastive learning-based approach to polyp matching using a Siamese neural network architecture. The model was trained on a dataset from the Danish CareForColon2015 trial. The dataset comprises 5,103 image pairs of confirmed same polyp and 5,103 pairs of different polyps, annotated by experienced clinicians. The proposed model employs ResNet variants (ResNet18, ResNet34, ResNet50), DenseNet121 and EfficientNet B0/B3 as backbones and utilizes contrastive learning with cosine vector distance metric for classification. Experimental results demonstrate high classification accuracies of up to 92\% and AUC values close to 0.98. These results suggest that automated polyp matching has the potential to reduce polyp over-counting, improve patient risk stratification, and reduce unnecessary follow-up colonoscopies.
\end{abstract}

\acresetall


\section{Background}
\label{sec:background}
\Ac{CCE} is increasingly employed for imaging and inspection of the colon~\citep{Spada2015, Friedel2016, Hosoe2021, Koulaouzidis2021, Moen2022, Altobelli2025, Lei2025a, Lei2025b, Altobelli2025_reply}. In a \ac{CCE} procedure, the patient swallows a capsule camera, which then traverses the gastrointestinal tract and exits via the anus. Prior to \ac{CCE}, the patient goes through a procedure to clean their intestines~\citep{Beltran2008, Spada2012}. \ac{CCE} lacks therapeutic capabilities, and when significant polyps are found in a \ac{CCE} video, their removal requires a follow-up colonoscopy~\citep{Jung2012}. However, if the clinician is inaccurately informed of a higher number of polyps than actually present, additional time during colonoscopy may be spent searching for non-existent polyps. Furthermore, since the number of polyps identified is part of the risk allocation the patients receive, double counting of polyps may ultimately result in unnecessary follow-up, causing additional costs and discomfort or anxiety to patients~\citep{Ghanouni2016, Koulaouzidis2021, BjorsumMeyer2022}.

There are several colon capsule/pill cameras on the market that are quite similar~\citep{Scallan2016, Gounella2023}. Data of the current study are derived from \ac{CCE} procedures conducted using second generation colon capsules (PillCam\texttrademark\ Colon 2, Medtronic, USA)~\citep{Negreanu2013}. This pill camera is \qty{31.5}{mm} long and has a diameter of \qty{11.6}{mm}. Each head of the pill is equipped with a camera that can function independently of the other. Each camera head has a view angle of \ang{172} to all sides, which in total mounts to almost \ang{360} view. Together, the two camera heads achieve a maximum frame rate of 35 frames per second. Typically, each camera head captures 5,000--10,000 images. 

In a clinical setting, the camera capsule is known to exhibit bidirectional movement, occasionally traveling backwards between different colonic segments. Consequently, polyps and other features may be visualized multiple times during a \ac{CCE} investigation~\citep{Buijs2018, BjorsumMeyer2022, Lei2024b}. To prevent the same polyp from being counted more than once, it is essential to have a reliable method for determining whether two or more images of polyps correspond to the same polyp. This task is called ``polyp matching'' and is essential in \ac{CCE}, and the integration of \ac{AI} into this process is crucial~\citep{Nadimi2025}.

There are several factors that make polyp matching based on solely images a challenging task~\citep{Lei2024a}. First, a single polyp can appear significantly different when viewed from varying angles or distances. Second, the capsule's onboard lighting can create diverse illumination conditions, including specular reflections. Third, the cleanliness of the colon can change over time affecting image consistency. In a setting where a clinician needs to determine whether two polyp identifications are of the same polyp, a list of relevant factors will be assessed. These include the time interval between images, the estimated location within the colon, the size and type of the polyps~\citep{Lei2024a, Lei2024b}. However, this is a challenging task where there will be made errors by clinicians, and especially by non-expert clinicians~\citep{Buijs2018}.

In this paper, we  focus on a fundamental question: ``Given two images, both assumed to contain a polyp, are they showing the same polyp?''. To the best of our knowledge, this is the first attempt towards addressing this research question for \ac{CCE}~\citep{Lei2024a}. Within conventional colonoscopy, the typical situation is a bit different where the scope first is inserted in the rectum and moved to the start of colon (cecum)~\citep{Rex2015}. Clinical practice varies slightly regarding what is done during this insertion process. Some clinicians remove (at least small) polyps detected during the insertion process, during which they do a less thorough colon inspection compared to during the scope retraction process. Other clinicians only remove detected polyps during the retraction process from start (cecum) to end of colon (anus). This approach risks missing polyps detected during the insertion process. As the scope is normally inserted fully and then retracted without larger back-and-forth movements in the colon, the clinician will normally not be unsure about whether a detected polyp was revisited (except this aspect of polyps detected during scope insertion). However, for conventional colonoscopy there has been work done on tracking polyps and identifying the correct number of polyps. \citet{Parolari2025a, Parolari2025b} use an architecture of ResNet and transformers to track, cluster and count polyps across consecutive frames. Similarly, \citet{Intrator2023} present a method for polyp re-identification in colonoscopy videos using self-supervised learning, combining short-term tracking with appearance-based re-identification. However, such an approach is not applicable to \ac{CCE} where images of the same polyp may be separated by several hours due to the capsule's passive and bidirectional movement, which makes tracking impossible.

The main contributions of this paper are: (i) the first deep learning framework for polyp matching in \ac{CCE}; (ii) a well-curated dataset of 5,103 same-polyp image pairs from the CareForColon2015 clinical trial; (iii) a Siamese network with pairwise contrastive loss achieving approximately 92\% accuracy and AUC of 0.98; and (iv) a comparative evaluation of six CNN backbone architectures (ResNet18/34/50, DenseNet121, EfficientNet B0/B3) confirming that performance is robust to backbone choice.

The paper is organized as follows: Section~\ref{sec:methodology} describes the dataset and methodology, Section~\ref{sec:results} presents the experimental results, and Sections~\ref{sec:discussion} and~\ref{sec:conclusion} provide discussion and conclusions.

\section{Methodology}
\label{sec:methodology}
This section introduces the dataset and the methodology of this study. 

\subsection{Dataset}
In order to address the research question, we need at least two images known to be of the same polyp. Ideally, one would like to have images separated in time of the same polyp and seen from multiple directions. However, no such curated public dataset exists. In this study, we used the Danish CareForColon2015 trial~\citep{Kaalby2020, Baatrup2025} as the main data source. From this dataset, clinicians from the Department of Surgery of Odense University Hospital have for a large set of polyps exported up to five images of the same polyp from the same camera pill passing (so there are no revisit images of polyps). The five exported images show 1. First partial, 2. First full, 3. Best full, 4. Last full, 5. Last partial, image of the same polyp. This means that 1 and 5 might just show parts of the polyp or the polyp might be very far away in the image, while 2--4 show the whole polyp. This did not give us images of the same polyp separated significantly in time. However, for each polyp we can use three pairs (2-3, 2-4, 3-4) of images showing the same full polyp. We used these pairs as training data for the ``same polyp'' class. For the ``different polyps'' class, random sets of images from different polyps are chosen, as described below.

In a real clinical setting, one would of course only use images of the same patient when doing inference. This can be done also in the training setting presented here, but this has at least two issues connected to it. Firstly, this restricts the analysis to patients with more than one polyp. This might mean that we undersample smaller polyps, which is expected to be on average the case for patients with a single polyp. Secondly, we know that there are some cases where the annotator has mislabeled images as being of different polyps, while they are really  of the same polyp.

We have for the current study access to 2,780 polyps from 853 patients (The full dataset is constantly growing and currently consists of about 6,000 polyps, but all these have not been exported to this five-image-format described above). From this data set, we only use polyps that have been assigned all five images (partial, full etc), to be sure that we are working with images that show the whole polyp (images 2--4). This results in 1,701 polyps from 722 patients. From 1,701 polyps, each with three pairs, we get 5,103 image pairs in the ``same polyp'' class. From the very large number of possible pairs of different polyp images, we drew a random subset of 5,103 pairs for the ``different polyp'' class. As mentioned above, as the same polyp might wrongly be labeled as different polyps for a given patient, we do not use images from assumed different polyps from the same patient. I.e., a pair of non-same-polyp images are from two different patients. In the end, we have a balanced training set of 5,103 sample pairs for each of the two classes.

\begin{figure}[htbp]
    \centering
    \includegraphics[width=\linewidth]{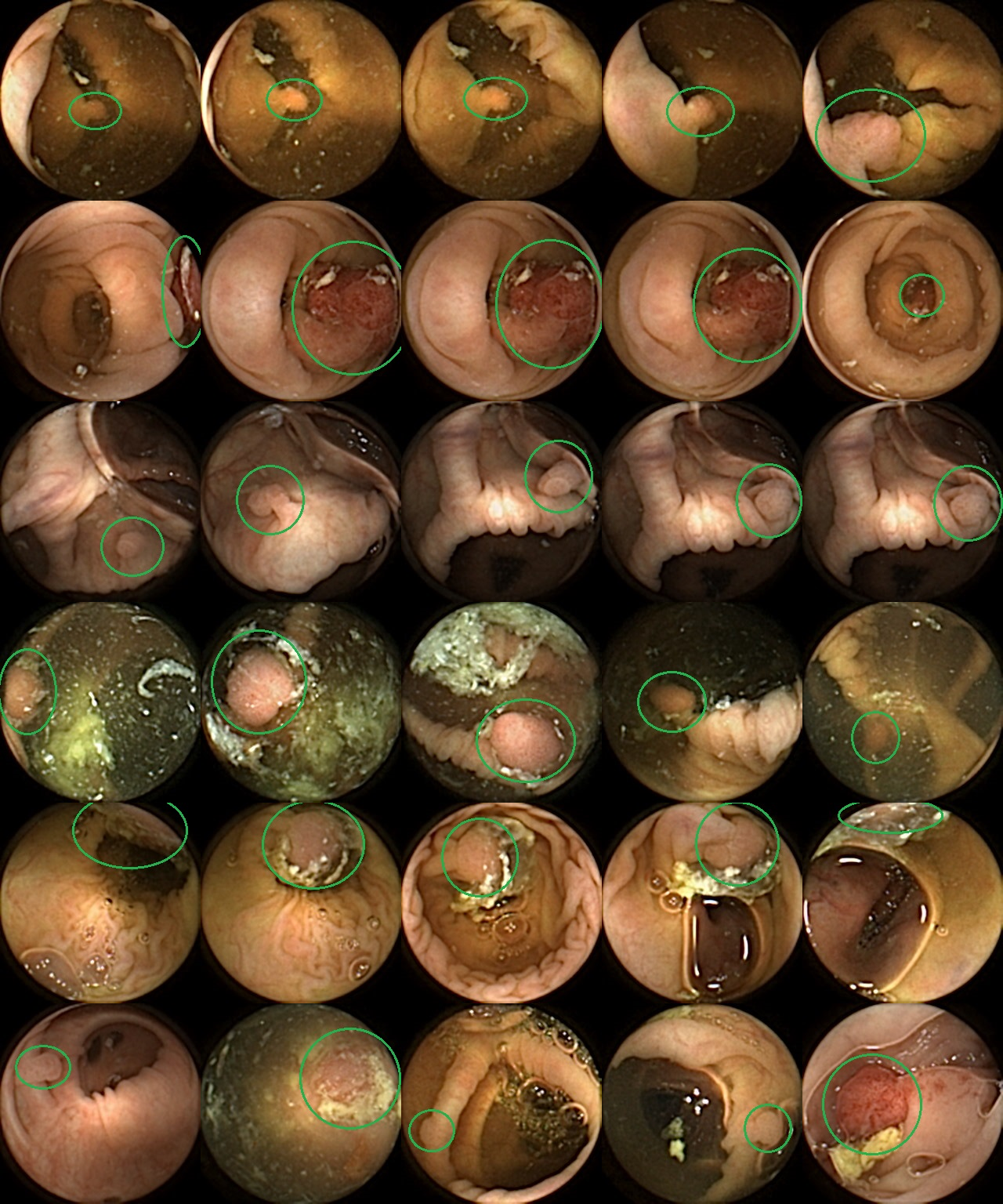}
    \caption{Rows 1--5: Each row shows images of a single polyp. Row 6: Example images of five different polyps. The figure also shows varying degree of cleansing quality, artifacts and partially visible polyps. The polyps are marked by a green ellipse.}
    \label{fig:Polyp_examples}
\end{figure}

Figure~\ref{fig:Polyp_examples} shows examples of these 1--5 images. The cleansing quality and presence of artifacts (like bubbles) can vary. We can note that depending on the angle or distance and the presence of artifacts in different images of the same polyp, the task of matching polyps can become very challenging.

All data splits were made at the patient level: all polyps, and hence all image pairs, of a given patient belong to exactly one of the training, validation or test sets, so there is no data leakage between sets. The splits are described in Section~\ref{sec:training} and summarized in Table~\ref{tab:dataset}.

\begin{table}[htbp]
    \centering
    \caption{Overview of the datasets used in this study. As all polyps of a given patient will remain in the same data set or fold split, the number of polyps within a fold will vary slightly. The set of ``Test pairs'' of ``Split 2'' is the same as the ``missing'' 15\% not used in the \acs{CV} of Split 1. 
}
    \label{tab:dataset}
    \begin{tabular}{@{} lcc @{}}
        \toprule
                   & Number & Fraction of total \\
         \midrule
         Patients in total & 722 & 1.00\\
         Polyps in total & 1701 & 1.00\\
         Polyp pairs & 5103 &  1.00\\
         Split 1: Cross-validation & & \\
         - Polyp pairs & 4338 & 0.85 \\
         Split 2: Final full model & &  \\
         - Training pairs & 3572 & 0.70 \\
         - Validation pairs & 766 & 0.15 \\
         - Test pairs & 765 & 0.15 \\
         \bottomrule
    \end{tabular}
\end{table}
Appendix~\ref{sec:app_A} documents the size and morphological distribution of this dataset. As the dataset is not intentionally filtered, it should resemble a general \ac{CCE} population. It also includes information about the typical transit times across a polyp.

\subsection{Network architecture}
In this section, we discuss the suggested Siamese deep neural network for comparing images.

\begin{figure}[htbp]
    \centering
    \includegraphics[width=\linewidth]{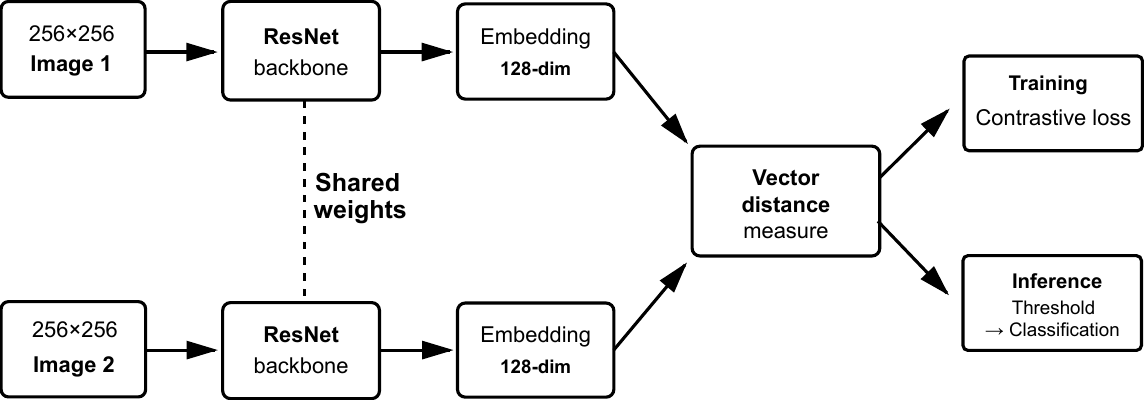}
    \caption{Network architecture of the proposed Siamese neural network with ResNet backbone and contrastive loss. Each input image (256x256 pixels, resized to 224x224 before processing) is independently fed through a shared backbone, producing a 128-dimensional embedding vector. During training, the cosine distance between the two embedding vectors is used to compute the contrastive loss, which pulls same-polyp pairs together and pushes different-polyp pairs apart. During inference, the cosine distance is compared against a learned threshold, where pairs below the threshold are classified as the same polyp and pairs above as different polyps.}
    \label{fig:Siamese}
\end{figure}

To assess whether two images, both containing polyps, are of the same polyp, we trained a Siamese deep neural network~\citep{Bromley1993} based on contrastive learning~\citep{Chopra2005,Chen2020,Ilina2022,Zhang2022}. Figure~\ref{fig:Siamese} shows the network architecture for both training and inference mode. Contrastive learning is a class of metric learning approaches where a model learns to produce an embedding space in which similar inputs are mapped close together and dissimilar inputs far apart. It is well-suited for problems defined in terms of pairwise similarity rather than membership in a fixed set of classes. A closely related alternative is triplet loss~\citep{Schroff2015}, which requires one image to be designated as an anchor and the remaining two as a positive and a negative. This introduces an asymmetry that does not fit the polyp matching task naturally, where neither image in a pair has a privileged role. The pairwise contrastive loss therefore maps more directly onto both the structure of the training data, which consists of labelled same/different image pairs, and the clinical task of making a binary judgment between two candidate images.

A Siamese network consists of two identical networks that have the same structure and shared weights. When doing inference, two separate images are fed through each of these two ``arms'' of the Siamese network. The outputs of the two networks are then fed into an embedding layer of size 128, which results in a final embedding vector of length 128 for each image. The distance between these two embedding vectors forms the basis of the final classification, where a large distance is an indication of the images not being of the same polyp. 

The ResNet family~\citep{He2016_resnet} was chosen as the main backbone due to its well-documented performance across medical imaging tasks, the availability of ImageNet-pretrained weights, and availability of different model sizes (number of residual blocks). In this study we used ResNet18, ResNet34 and ResNet50 as the backbone.

The cosine vector distance $D_W^{\mathrm{cos}}(\bm{A}, \bm{B})$ was employed for measuring vector distances between embedding vectors $\bm{A}$ and $\bm{B}$. This is defined as 1 minus the cosine similarity $\cos(\bm{A}, \bm{B})$, which gives
\begin{equation}
    D_W^{\mathrm{cos}}(\bm{A}, \bm{B}) = 1 - \cos(\bm{A}, \bm{B}) = 1 - \frac{\bm{A} \cdot \bm{B}}{\norm{\bm{A}}\,\norm{\bm{B}}},
\end{equation}
where $\cdot$ is the vector dot product and $\norm{\cdot}$ is the L2 norm.

We used a contrastive loss function~\citep{Hadsell05}. This attempts to push images of different polyps further apart, while images of the same polyp should have as low as possible vector distance~\citep{Xiao2021}. This loss is defined as
\begin{equation}
    L_{C} = (1-Y)\cdot D_W^2 + Y\cdot \max(0, m-D_W)^2, 
\end{equation}
where $Y$ is 0/1 for same/different polyp image pairs and $D_W$ is some distance measure between the two embedding vectors. The margin $m$ is there to stop the network from focusing on already large vector differences for images pairs of different polyps. That is, if the vector distance is larger than $m$ for a pair of different polyp images, this pair will not contribute to the loss function. We employed the commonly chosen cosine vector distance margin of $m=0.5$, which corresponds to cosine similarity of 0.5, which again corresponds to an angular separation of \ang{60}.

\subsection{Training}
\label{sec:training}
In this section, we present the training approaches and parameters of our experiments. 

The raw input images are 256x256 pixels. Even though ResNets can take many input dimensions, we resized the images to 224x224 pixels to fit the input dimensions of the pretrained ResNets (PyTorch with weights IMAGENET1K\_V1) we used. As the resizing factor is small, we do not expect to lose any significant detail level. The images were scaled to have zero mean and 1 standard deviation for each color channel. We used on-the-fly data augmentation techniques~\citep{Yang2023} to increase the effective sample size. We used random rotation (all angles), random horizontal and vertical flips (p=0.5), and color jittering (contrast=brightness=0.2).

In all experiments we ran the training for up to 100 epochs and enforced early stopping with a patience value of 20 and minimum loss decrease of 0.001 to avoid overfitting. This means that if a model trained for 40 epochs before early stopping is triggered, the model state at 20 epochs was chosen as the best model.

During the training phase, the threshold value for the cosine distance was set to 0.15. After the training had completed or early stopping had been triggered, the optimal threshold value was estimated for each fold. This was done via running a grid search over possible threshold values on the validation set, and keeping the threshold value that maximized accuracy. 

As both classes are equally represented in the validation set, maximizing accuracy is equivalent to maximizing the Youden index and will normally give a balanced confusion matrix, making explicit ROC-based threshold selection redundant. This optimal validation-set-based threshold was then used for doing inference on the test set.

Prior to any \ac{CV} training, we split off a set of patients and their polyps into a test set. This random test set constitutes 15\% of the total number of polyps (which also corresponds to approximately 15\% of the patients), and was not used in any way until the final performance evaluation. The remaining 85\% was used as the ``CV data set''. Different-polyp pairs were only formed from patients within the same set.

We ran 10-fold \ac{CV} where we ran 10 fully independent training runs. For each of the 10 folds, we had a test set of 10\% of the \acs{CV} data set, and the remaining 90\% was split into 80\% training and 20\% validation. From the results of the 10 folds, we report mean and standard deviations of performance measures, and for each fold there is an estimated optimal threshold parameter that is recorded.

Following the \ac{CV}, we performed a final training run on the whole dataset. Here we used the earlier split off 15\% test set as the final test set, while the training and validation sets were 70\% and 15\% of the full data set, respectively. The validation set was used for early stopping and threshold selection. The performance measures of this final test set are reported as the final results of this study. We used the same other training parameters as for the \ac{CV} training.

The networks were trained on The University of Southern Denmark's Ucloud cluster solution on a NVIDIA H100 with 192 GB RAM. We used the PyTorch framework via NVIDIA's optimized container release 25.11. Table~\ref{tab:training_params} shows the training parameters used in this study.
\begin{table}[htbp]
    \centering
    \caption{Parameters used in Siamese deep neural net training. The ``train/val split'' for \acs{CV} is within the 90\% of the \acs{CV} data set used in that fold. The remaining 10\% of each fold is used to evaluate the test set performance of that \acs{CV} split.}
    \label{tab:training_params}
    \begin{tabular}{@{} lc @{\hspace{0.5cm}} lc @{}}
        \toprule
         Parameter          & Value             &  Parameter                    & Value  \\
         \midrule
         Backbones          & ResNet 18/34/50   & Image resize                  & 256x256 to 224x224 \\
         Max epochs         & 100               & Augmentation                  & On-the-fly \\
         Batch size         & 64                & - Random rotation             & Uniform, \ang{0}--\ang{360} \\
         Early stopping     &                   & - Vertical/horizontal flip    & p=0.5 \\
         - Patience         & 20                & - Contrast jitter             & 0.2\\
         - min loss decrease& 0.001             & - Brightness jitter           & 0.2\\
         Loss               & Contrastive       & Cross-validation              & \\
         Optimizer          & Adam              & - Folds                       & 10 \\ 
         - Learning rate    & 0.0001            & - Train/val split             & 80/20 \\
         - Weight decay     & 0.0001            &                               &  \\
         \bottomrule
    \end{tabular}
\end{table}

\section{Results}
\label{sec:results}
This section presents the results of the 10-fold \ac{CV} training and the results of the final models based on the full dataset. 

For the \acs{CV} analysis, we present the estimated mean of the performance measures and their estimated standard deviation for all three backbones. As described above, the confusion matrices are expected to on average be symmetric, so we do not present them here. Each training run was at max run for 100 epochs and early stopping was enforced (and occurred in all \ac{CV} runs). As we have run 10 folds and three ResNets backbones, the \acs{CV} part consisted of 30 trained models.  Table~\ref{tab:performance_CV} shows the held-out test set results of various performance measures for the different ResNet backbones.

\begin{table}[htbp]
\centering
\caption{Cross-validation test set performance measures for different ResNets. Mean values and $\pm$ 1 standard deviation across the 10 folds. AUC is area under the receiver operating characteristic curve.}
\label{tab:performance_CV}
\begin{tabular}{@{} l ccc @{}}
\toprule
 Metric             & ResNet18            & ResNet34  & ResNet50 \\
 \midrule
Accuracy            & 0.9243 $\pm$ 0.0136 & 0.9231 $\pm$ 0.0096 & 0.9167 $\pm$ 0.0085 \\
Precision           & 0.9253 $\pm$ 0.0179 & 0.9181 $\pm$ 0.0131 & 0.9148 $\pm$ 0.0168 \\
Recall              & 0.9234 $\pm$ 0.0148 & 0.9292 $\pm$ 0.0125 & 0.9194 $\pm$ 0.0112 \\
Specificity         & 0.9252 $\pm$ 0.0191 & 0.9170 $\pm$ 0.0143 & 0.9140 $\pm$ 0.0195 \\
F1 score            & 0.9242 $\pm$ 0.0135 & 0.9236 $\pm$ 0.0095 & 0.9170 $\pm$ 0.0079 \\
AUC                 & 0.9780 $\pm$ 0.0051 & 0.9775 $\pm$ 0.0056 & 0.9759 $\pm$ 0.0053 \\
Optimal threshold   & 0.2360 $\pm$ 0.0150 & 0.2190 $\pm$ 0.0104 & 0.2170 $\pm$ 0.0142 \\
Early stopped epoch & 12.7   $\pm$ \phantom{0}5.9    & 11.1   $\pm$ \phantom{0}3.9    & 10.1   $\pm$ \phantom{0}6.2 \\
\bottomrule
\end{tabular}
\end{table}

\begin{figure}[htbp]
    \centering
    \includegraphics[width=1\linewidth]{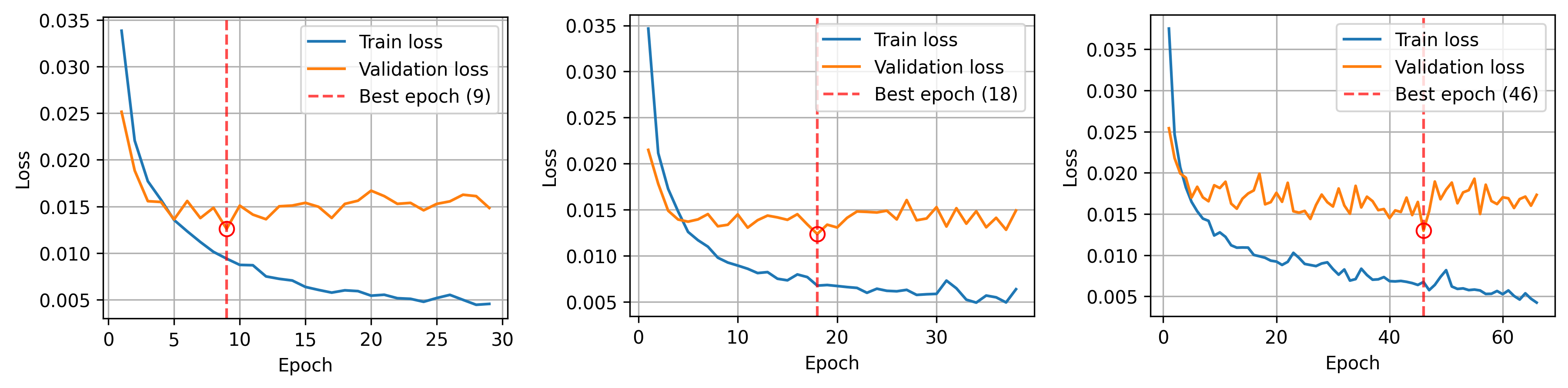}
    \caption{Training and validation loss using cosine distance metrics, and ResNet of size 18, 34 and 50 from left to right. Best model following triggering of early stopping marked in red.}
    \label{fig:training_curve_final}
\end{figure}

The final test set consists of 113 patients with 255 polyps. As each polyp creates three same-polyp pairs, there are $255\cdot3 = 765$ same-polyp pairs in the final test set (and the same number of different-polyp pairs). Table~\ref{tab:performance_final} shows the performance measures of the final models. Figure~\ref{fig:training_curve_final} shows the final training curves, with the best model marked (following early stopping). It can be seen that early stopping was triggered for all three.

\begin{table}[htbp]
\centering
\caption{Test set performance measures for different ResNets. AUC is area under the receiver operating characteristic curve.}
\label{tab:performance_final}
\begin{tabular}{@{} l ccc @{}}
\toprule
Metric & ResNet18 & ResNet34 & ResNet50 \\
\midrule
Accuracy & 0.9193 & 0.9225 & 0.9206 \\
Precision & 0.9237 & 0.9275 & 0.9184 \\
Recall & 0.9141 & 0.9167 & 0.9232 \\
Specificity & 0.9245 & 0.9284 & 0.9180 \\
F1 score & 0.9188 & 0.9221 & 0.9208 \\
AUC & 0.9754 & 0.9768 & 0.9804 \\
Optimal threshold & 0.25 & 0.23 & 0.22 \\
Early stopped epoch & 9 & 18 & 46\\
\bottomrule
\end{tabular}
\end{table}

\begin{figure}[htbp]
    \centering
    \includegraphics[width=1\linewidth]{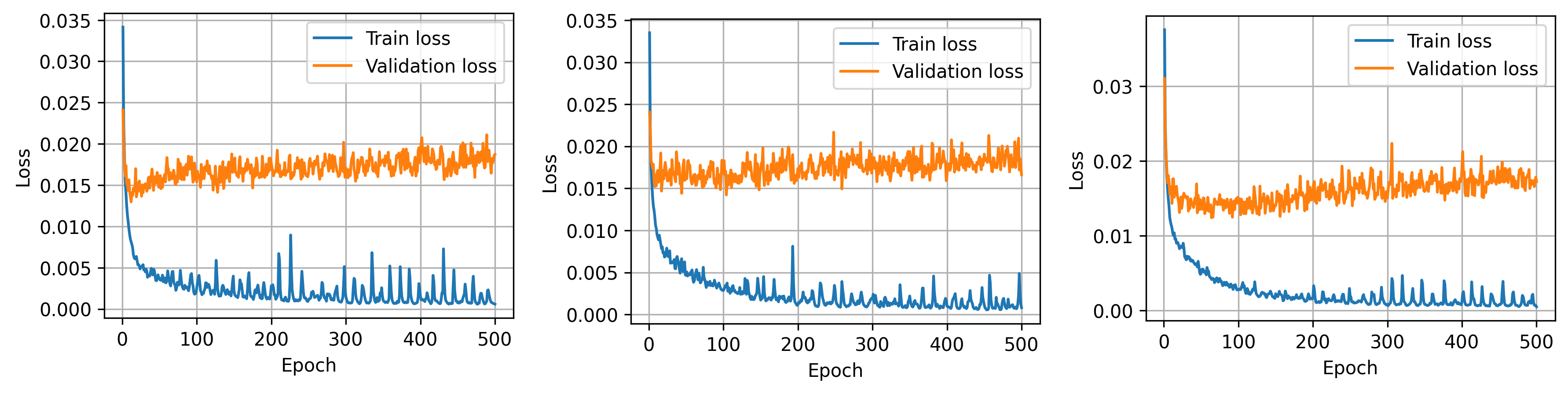}
    \caption{Training and validation loss when  running for 500 epochs. The other hyperparameters are as in Table~\ref{tab:training_params}, except that no early stopping is used. ResNet of size 18, 34 and 50 are shown from left to right, respectively.}
    \label{fig:training_curve_final_500}
\end{figure}

Even though the final models of this paper are selected following early stopping, we repeated the training processes without early stopping and let them run for 500 epochs. This was to make sure we did not miss an unexpected, high-epoch loss decrease. Figure~\ref{fig:training_curve_final_500} shows these 500-epoch training curves, and it can be seen that all curves have an increasing validation loss as the epoch number gets high. This does not mean that other training setups might not produce better results.

The runtime varied depending on the size of the ResNet used, with mean values of 9.4, 11.8 and 15.3 seconds/epoch for ResNet18, ResNet34 and ResNet50, respectively. As early stopping occurred for all runs, the actual training time for the \ac{CV} was 2 hours and 54 minutes to train 10-fold \ac{CV} of three ResNet backbones. The training time for the final models was 25 minutes for the three ResNet backbones.

To test whether other well-established models perform better than the ResNets, we ran the full analysis also using DenseNet121~\citep{DenseNet2017}, EfficientNet~\citep{EfficientNet2019} B0 and B3 as backbones. The results of these analyses are found in Appendix~\ref{sec:app_B}. As can be seen, the differences in results seem to be so small that it is hard to claim that one backbone is clearly better than the rest.

Appendix~\ref{sec:app_C} displays the cosine vector distance distribution for the various set-ups and backbones. 

\section{Discussion}
\label{sec:discussion}

This study presents the first deep learning-based approach to polyp matching in \ac{CCE}, achieving classification accuracy of up to 92.25\% (ResNet34) and AUC values up to 0.9804 (ResNet50). These results compare favorably with clinician inter-observer agreement on polyp detection in \ac{CCE}, which has been reported to be imperfect even among experts~\citep{Buijs2018}, suggesting that automated matching support could meaningfully reduce errors in clinical reading.

Across the six backbone architectures evaluated, performance was consistently high, with the ResNets slightly outperforming EfficientNet and DenseNet variants. The similarity in performance across architectures suggests that the contrastive learning framework itself, rather than any specific backbone, is the primary driver of the results, and that the framework generalizes across CNN architectures.

The intended clinical use of this tool is as a decision support component within a \ac{CCE} analysis pipeline. After a polyp detection algorithm flags candidate polyp images~\citep{Nadimi2025}, the matching algorithm can be applied to any pair of images, outputting a binary same/different classification. Relevant questions in a clinical deployment would be: should all possible polyp image pairs be compared, even those separated by only a very short time? If not all pairs, how does one select which pairs to compare? In practice, this algorithm would serve as a supporting tool alongside other information such as time difference, polyp size, polyp type, and estimated position in the colon. Running inference on a pair of images typically takes about 40 milliseconds per compared pair on a NVIDIA RTX 2000 Ada Generation Laptop GPU. This means that it is feasible to compare a large number of polyp pairs even on normal GPUs, which facilitates integration of this matching procedure into existing workflows.

The main strength of the study is that we have a large set of images known to be of the same polyp, from which we have created a large and well-curated dataset. Additional strengths are that the strict patient-level data splitting ensures that no patient contributes to more than one of the train, validation, or test sets, and that the consistent performance across six different backbone architectures suggests that the results are not an artifact of any particular model choice. 

However, there are important limitations to acknowledge. Most importantly, the presented setup does not include pairs of images known to be of the same polyp but separated significantly in time, which will typically be the case in a real clinical setting. A typical polyp image pair to be compared in clinical use is therefore expected to on average be less similar than what is shown in Figure~\ref{fig:Polyp_examples}, due to changes in illumination, viewpoint, and bowel preparation over time. As a consequence, the presented performance numbers are likely to be overestimates compared to real-world clinical performance. We have also only included images that show the whole polyp, which will not always be the case in practice. Furthermore, all different-polyp pairs were drawn from two different patients. In a real clinical setting the compared polyps will always be from the same patient. Within-patient pairs share to some extent bowel cleansing quality and mucosal appearance, and are therefore likely harder to separate, which may also contribute to an overestimation of performance. Finally, we only included polyps for which the operator was able to export five images. Whether this introduces any bias into the dataset is not clear, but as the same-polyp and different-polyp pairs are constructed from the same set of polyps, this is unlikely to be a significant issue for clinical use.

\section{Conclusion}
\label{sec:conclusion}

We propose a novel deep learning framework that addresses a critical challenge in \ac{CCE}. The presented neural network takes two \ac{CCE} images known to both contain polyps and tries to determine whether they are of the same polyp. Having such an automated tool available when evaluating the results of a~\ac{CCE} investigation would be helpful to avoid over-counting the number of detected polyps. This will both ensure more accurate patient risk allocation and ultimately correct patient follow-up and save procedure time when a clinician at a later time goes in and removes polyps.

Future work will focus on creating a dataset of images of the same polyps, but clearly separated in time. Furthermore, we plan to apply explainable \ac{AI} methods to identify what is causing the wrong classifications of the presented method. It is also of interest to quantify how cleaning/image quality affects classification performance. Finally, it seems natural to expand from comparing two single images to comparing two sets of (neighboring) polyp images.

The presented approach is a first-of-a-kind feasibility study that shows the potential of using deep neural networks for polyp matching in colon capsule endoscopy, with the potential to save time and improve care in colon capsule endoscopy.


\subsection*{Ethics declaration}
The data used in this paper is from the CareForColon2015 trial. This was approved by The Regional Health Research Ethics Committee (Registration number: S-20190100), the Danish data protection agency (Ref. 19/29858), and has ClinicalTrials.gov registration number NCT04049357. All patients provided written consent to participate in the study. All work was done in accordance with the Declaration of Helsinki. 

\subsection*{Funding}
All authors were funded by the AICE project (101057400), which is a HORIZON EUROPE project funded by the European Union.

\subsection*{Availability of data and materials}
Data access requests should be made to Prof. Gunnar Baatrup at Odense University Hospital (OUH).

\subsection*{Acknowledgements}
This research is part of AICE project (number 101057400) funded by the European Union, and it is part-funded by the United Kingdom government. Views and opinions expressed are however those of the author(s) only and do not necessarily reflect those of the European Union or the European Commission. Neither the European Union nor the European Commission can be held responsible for them.
\begin{center}
    \includegraphics[height=1.1cm]{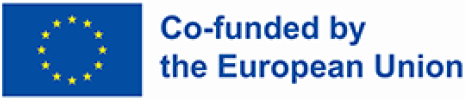}
    \includegraphics[height=1.1cm]{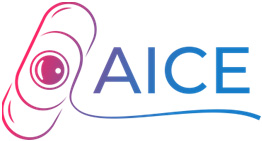}
\end{center}

\bibliographystyle{unsrtnat} 
\bibliography{AI_polyp_matching} 

\begin{thebibliography}{39}
\providecommand{\natexlab}[1]{#1}
\providecommand{\url}[1]{\texttt{#1}}
\expandafter\ifx\csname urlstyle\endcsname\relax
  \providecommand{\doi}[1]{doi: #1}\else
  \providecommand{\doi}{doi: \begingroup \urlstyle{rm}\Url}\fi

\bibitem[Spada et~al.(2015)Spada, Hassan, Campanale, and Costamagna]{Spada2015}
Cristiano Spada, Cesare Hassan, Mariachiara Campanale, and Guido Costamagna.
\newblock Colon capsule endoscopy.
\newblock \emph{Techniques in Gastrointestinal Endoscopy}, 17\penalty0
  (1):\penalty0 19--23, 2015.
\newblock \doi{10.1016/j.tgie.2015.02.005}.

\bibitem[Friedel et~al.(2016)Friedel, Modayil, and Stavropoulos]{Friedel2016}
David Friedel, Rani Modayil, and Stavros Stavropoulos.
\newblock Colon capsule endoscopy: Review and perspectives.
\newblock \emph{Gastroenterology Research and Practice}, pages 1--6, 2016.
\newblock \doi{10.1155/2016/9643162}.

\bibitem[Hosoe et~al.(2021)Hosoe, Limpias~Kamiya, Hayashi, Sujino, Ogata, and
  Kanai]{Hosoe2021}
Naoki Hosoe, Kenji J.~L. Limpias~Kamiya, Yukie Hayashi, Tomohisa Sujino,
  Haruhiko Ogata, and Takanori Kanai.
\newblock Current status of colon capsule endoscopy.
\newblock \emph{Digestive Endoscopy}, 33\penalty0 (4):\penalty0 529--537, 2021.
\newblock \doi{10.1111/den.13769}.

\bibitem[Koulaouzidis et~al.(2021)Koulaouzidis, Dabos, Philipper, Toth, and
  Keuchel]{Koulaouzidis2021}
Anastasios Koulaouzidis, Konstantinos Dabos, Michael Philipper, Ervin Toth, and
  Martin Keuchel.
\newblock How should we do colon capsule endoscopy reading: a practical guide.
\newblock \emph{Therapeutic Advances in Gastrointestinal Endoscopy},
  14:\penalty0 26317745211001983, 2021.
\newblock \doi{10.1177/26317745211001983}.

\bibitem[Moen et~al.(2022)Moen, Vuik, Kuipers, and Spaander]{Moen2022}
Sarah Moen, Fanny E.~R. Vuik, Ernst~J. Kuipers, and Manon C.~W. Spaander.
\newblock Artificial intelligence in colon capsule endoscopy—a systematic
  review.
\newblock \emph{Diagnostics}, 12\penalty0 (8):\penalty0 1994, 2022.
\newblock \doi{10.3390/diagnostics12081994}.

\bibitem[Altobelli et~al.(2025{\natexlab{a}})Altobelli, Angeletti, Varesini,
  Bianchi, and Masedu]{Altobelli2025}
Emma Altobelli, Paolo~Matteo Angeletti, Paolo~Angelo Varesini, Zuleyka Bianchi,
  and Francesco Masedu.
\newblock Colon capsule endoscopy as a promising diagnostic tool in colorectal
  cancer: A systematic review and network meta-analysis.
\newblock \emph{Diagnostics}, 15\penalty0 (17):\penalty0 2157,
  2025{\natexlab{a}}.
\newblock \doi{10.3390/diagnostics15172157}.

\bibitem[Lei et~al.(2025{\natexlab{a}})Lei, Valdivia, Marlicz,
  Skonieczna-\.{Z}ydecka, Arasaradnam, Eliakim, and Koulaouzidis]{Lei2025a}
Ian~Io Lei, Pablo~Cortegoso Valdivia, Wojciech Marlicz, Karolina
  Skonieczna-\.{Z}ydecka, Ramesh Arasaradnam, Rami Eliakim, and Anastasios
  Koulaouzidis.
\newblock Systematic meta-review: diagnostic accuracy of colon capsule
  endoscopy for colonic neoplasia with umbrella meta-analysis.
\newblock \emph{Therapeutic Advances in Gastrointestinal Endoscopy},
  18:\penalty0 26317745251370845, 2025{\natexlab{a}}.
\newblock \doi{10.1177/26317745251370845}.

\bibitem[Lei et~al.(2025{\natexlab{b}})Lei, Arasaradnam, Marlicz, and
  Koulaouzidis]{Lei2025b}
Ian~Io Lei, Ramesh Arasaradnam, Wojciech Marlicz, and Anastasios Koulaouzidis.
\newblock Comment on {Altobelli} et al. colon capsule endoscopy as a promising
  diagnostic tool in colorectal cancer: A systematic review and network
  meta-analysis. {Diagnostics} 2025, 15, 2157.
\newblock \emph{Diagnostics}, 15\penalty0 (23):\penalty0 2938,
  2025{\natexlab{b}}.
\newblock \doi{10.3390/diagnostics15232938}.

\bibitem[Altobelli et~al.(2025{\natexlab{b}})Altobelli, Angeletti, and
  Masedu]{Altobelli2025_reply}
Emma Altobelli, Paolo~Matteo Angeletti, and Francesco Masedu.
\newblock Reply to {Lei} et al. comment on {``Altobelli} et al. colon capsule
  endoscopy as a promising diagnostic tool in colorectal cancer: A systematic
  review and network meta-analysis. {Diagnostics} 2025, 15, 2157''.
\newblock \emph{Diagnostics}, 15\penalty0 (23):\penalty0 2939,
  2025{\natexlab{b}}.
\newblock \doi{10.3390/diagnostics15232939}.

\bibitem[Pons~Beltrán et~al.(2008)Pons~Beltrán, Carretero, González-Suárez,
  Fernandez-Urien, and Muñoz-Navas]{Beltran2008}
Vicente Pons~Beltrán, Cristina Carretero, Begoña González-Suárez, Iñaqui
  Fernandez-Urien, and Miguel Muñoz-Navas.
\newblock Intestinal preparation prior to capsule endoscopy administration.
\newblock \emph{World journal of gastroenterology}, 14\penalty0 (37):\penalty0
  5773--5775, 2008.
\newblock \doi{10.3748/wjg.14.5773}.

\bibitem[Spada et~al.(2012)Spada, Hassan, Galmiche, Neuhaus, Dumonceau, Adler,
  Epstein, Gay, Pennazio, Rex, Benamouzig, de~Franchis, Delvaux, Devière,
  Eliakim, Fraser, Hagenmuller, Herrerias, Keuchel, Macrae, Munoz-Navas,
  Ponchon, Quintero, Riccioni, Rondonotti, Marmo, Sung, Tajiri, Toth,
  Triantafyllou, Gossum, and Costamagna]{Spada2012}
C.~Spada, C.~Hassan, J.P. Galmiche, H.~Neuhaus, J.M. Dumonceau, S.~Adler,
  O.~Epstein, G.~Gay, M.~Pennazio, D.K. Rex, R.~Benamouzig, R.~de~Franchis,
  M.~Delvaux, J.~Devière, R.~Eliakim, C.~Fraser, F.~Hagenmuller, J.M.
  Herrerias, M.~Keuchel, F.~Macrae, M.~Munoz-Navas, T.~Ponchon, E.~Quintero,
  M.E. Riccioni, E.~Rondonotti, R.~Marmo, J.J. Sung, H.~Tajiri, E.~Toth,
  K.~Triantafyllou, A.~Van Gossum, and G.~Costamagna.
\newblock Colon capsule endoscopy: {European Society of Gastrointestinal
  Endoscopy} ({ESGE}) guideline.
\newblock \emph{Endoscopy}, 44\penalty0 (5):\penalty0 527--536, 2012.
\newblock \doi{10.1055/s-0031-1291717}.

\bibitem[Jung(2012)]{Jung2012}
M.~Jung.
\newblock The ``difficult'' polyp: Pitfalls for endoscopic removal.
\newblock \emph{Digestive Diseases}, 30\penalty0 (Suppl. 2):\penalty0 74--80,
  2012.
\newblock \doi{10.1159/000341898}.

\bibitem[Ghanouni et~al.(2016)Ghanouni, Plumb, Hewitson, Nickerson, Rees, and
  von Wagner]{Ghanouni2016}
Alex Ghanouni, Andrew Plumb, Paul Hewitson, Claire Nickerson, Colin~J. Rees,
  and Christian von Wagner.
\newblock Patients' experience of colonoscopy in the {English Bowel Cancer
  Screening Programme}.
\newblock \emph{Endoscopy}, 48\penalty0 (3):\penalty0 232--240, 2016.
\newblock \doi{10.1055/s-0042-100613}.

\bibitem[Bjørsum-Meyer et~al.(2022)Bjørsum-Meyer, Koulaouzidis, and
  Baatrup]{BjorsumMeyer2022}
Thomas Bjørsum-Meyer, Anastasios Koulaouzidis, and Gunnar Baatrup.
\newblock The optimal use of colon capsule endoscopes in clinical practice.
\newblock \emph{Therapeutic Advances in Chronic Disease}, 13:\penalty0
  20406223221137501, 2022.
\newblock \doi{10.1177/20406223221137501}.

\bibitem[Scallan(2016)]{Scallan2016}
R.~Scallan.
\newblock State of the art inside view, the camera pill.
\newblock \emph{Technology and Health Care}, 24\penalty0 (4):\penalty0
  471--481, 2016.
\newblock \doi{10.3233/THC-161149}.

\bibitem[Gounella et~al.(2023)Gounella, Granado, Hideo Ando~Junior, Luporini,
  Gazziro, and Carmo]{Gounella2023}
Rodrigo Gounella, Talita~Conte Granado, Oswaldo Hideo Ando~Junior, Daniel~Luís
  Luporini, Mario Gazziro, and João~Paulo Carmo.
\newblock Endoscope capsules: The present situation and future outlooks.
\newblock \emph{Bioengineering}, 10\penalty0 (12):\penalty0 1347, 2023.
\newblock \doi{10.3390/bioengineering10121347}.

\bibitem[Negreanu et~al.(2013)Negreanu, Babiuc, Bengus, and
  Sadagurschi]{Negreanu2013}
Lucian Negreanu, Ruxandra Babiuc, Andreea Bengus, and Roxana Sadagurschi.
\newblock Pillcam {Colon} 2 capsule in patients unable or unwilling to undergo
  colonoscopy.
\newblock \emph{World journal of gastrointestinal endoscopy}, 5\penalty0
  (11):\penalty0 559--567, 2013.
\newblock \doi{10.4253/wjge.v5.i11.559}.

\bibitem[Buijs et~al.(2018)Buijs, Kroijer, Kobaek-Larsen, Spada,
  Fernandez-Urien, Steele, and Baatrup]{Buijs2018}
Maria~Magdalena Buijs, Rasmus Kroijer, Morten Kobaek-Larsen, Cristiano Spada,
  Ignacio Fernandez-Urien, Robert~JC Steele, and Gunnar Baatrup.
\newblock Intra and inter-observer agreement on polyp detection in colon
  capsule endoscopy evaluations.
\newblock \emph{United European Gastroenterology Journal}, 6\penalty0
  (10):\penalty0 1563--1568, 2018.
\newblock \doi{10.1177/2050640618798182}.

\bibitem[Lei et~al.(2024{\natexlab{a}})Lei, Koulaouzidis, Baatrup, Samaan,
  Parisi, McAlindon, Toth, Shaukat, Valentiner, Dabos, Fernandez, Robertson,
  Schelde-Olesen, Parsons, and Arasaradnam]{Lei2024b}
Ian~Io Lei, Anastasios Koulaouzidis, Gunnar Baatrup, Mark Samaan, Ioanna
  Parisi, Mark McAlindon, Ervin Toth, Aasma Shaukat, Ursula Valentiner,
  Konstantinos~John Dabos, Ignacio Fernandez, Alexander Robertson, Benedicte
  Schelde-Olesen, Nicholas Parsons, and Ramesh~P. Arasaradnam.
\newblock Rationalizing polyp matching criteria in colon capsule endoscopy: an
  international expert consensus through {RAND} (modified {DELPHI}) process.
\newblock \emph{Therapeutic Advances in Gastroenterology}, 17:\penalty0 1--12,
  2024{\natexlab{a}}.
\newblock \doi{10.1177/17562848241242681}.

\bibitem[Nadimi et~al.(2025)Nadimi, Braun, Schelde-Olesen, Khare, Gogineni,
  Blanes-Vidal, and Baatrup]{Nadimi2025}
Esmaeil~S. Nadimi, Jan-Matthias Braun, Benedicte Schelde-Olesen, Smith Khare,
  Vinay~C. Gogineni, Victoria Blanes-Vidal, and Gunnar Baatrup.
\newblock Towards full integration of explainable artificial intelligence in
  colon capsule endoscopy’s pathway.
\newblock \emph{Scientific Reports}, 15:\penalty0 5960, 2025.
\newblock \doi{10.1038/s41598-025-89648-z}.

\bibitem[Lei et~al.(2024{\natexlab{b}})Lei, Arasaradnam, and
  Koulaouzidis]{Lei2024a}
Ian~Io Lei, Ramesh Arasaradnam, and Anastasios Koulaouzidis.
\newblock Polyp matching in colon capsule endoscopy: Pioneering
  {CCE}-colonoscopy integration towards an {AI}-driven future.
\newblock \emph{Journal of Clinical Medicine}, 13\penalty0 (23):\penalty0 7034,
  2024{\natexlab{b}}.
\newblock \doi{10.3390/jcm13237034}.

\bibitem[Rex et~al.(2015)Rex, Schoenfeld, Cohen, Pike, Adler, Fennerty, 2nd,
  Park, Rizk, Sawhney, Shaheen, Wani, and Weinberg]{Rex2015}
Douglas~K Rex, Philip~S Schoenfeld, Jonathan Cohen, Irving~M Pike, Douglas~G
  Adler, M~Brian Fennerty, John G~Lieb 2nd, Walter~G Park, Maged~K Rizk,
  Mandeep~S Sawhney, Nicholas~J Shaheen, Sachin Wani, and David~S Weinberg.
\newblock Quality indicators for colonoscopy.
\newblock \emph{American Journal of Gastroenterology}, 110\penalty0
  (1):\penalty0 72--90, 2015.
\newblock \doi{10.1038/ajg.2014.385}.

\bibitem[Parolari et~al.(2026)Parolari, Cherubini, Ballan, and
  Biffi]{Parolari2025a}
Luca Parolari, Andrea Cherubini, Lamberto Ballan, and Carlo Biffi.
\newblock Temporally-aware supervised contrastive learning for polyp counting
  in colonoscopy.
\newblock In \emph{Medical Image Computing and Computer Assisted Intervention -
  {MICCAI} 2025}, pages 543--553, 2026.
\newblock \doi{10.1007/978-3-032-05127-1_52}.

\bibitem[Parolari et~al.(2025)Parolari, Cherubini, Ballan, and
  Biffi]{Parolari2025b}
Luca Parolari, Andrea Cherubini, Lamberto Ballan, and Carlo Biffi.
\newblock Towards polyp counting in full-procedure colonoscopy videos.
\newblock In \emph{2025 IEEE 22nd International Symposium on Biomedical Imaging
  (ISBI)}, pages 1--5, 2025.
\newblock \doi{10.1109/ISBI60581.2025.10981023}.

\bibitem[Intrator et~al.(2023)Intrator, Aizenberg, Livne, Rivlin, and
  Goldenberg]{Intrator2023}
Yotam Intrator, Natalie Aizenberg, Amir Livne, Ehud Rivlin, and Roman
  Goldenberg.
\newblock Self-supervised polyp re-identification in colonoscopy.
\newblock In \emph{Medical Image Computing and Computer Assisted Intervention -
  {MICCAI} 2023}, pages 590--600, 2023.
\newblock \doi{10.1007/978-3-031-43904-9_57}.

\bibitem[Kaalby et~al.(2020)Kaalby, Deding, Kobaek-Larsen, Havshoi,
  Zimmermann-Nielsen, Thygesen, Kroijer, Bjørsum-Meyer, and
  Baatrup]{Kaalby2020}
Lasse Kaalby, Ulrik Deding, Morten Kobaek-Larsen, Anne-Line~Volden Havshoi,
  Erik Zimmermann-Nielsen, Marianne~Kirstine Thygesen, Rasmus Kroijer, Thomas
  Bjørsum-Meyer, and Gunnar Baatrup.
\newblock Colon capsule endoscopy in colorectal cancer screening: a randomised
  controlled trial.
\newblock \emph{BMJ Open Gastroenterology}, 7\penalty0 (1):\penalty0 e000411,
  2020.
\newblock \doi{10.1136/bmjgast-2020-000411}.

\bibitem[Baatrup et~al.(2025)Baatrup, Bjørsum-Meyer, Kaalby, Schelde-Olesen,
  Kobaek-Larsen, Koulaouzidis, Kroijer, Al-Najami, Buch, Høgh, Qvist,
  Thygesen, and Deding]{Baatrup2025}
G~Baatrup, T~Bjørsum-Meyer, L~Kaalby, B~Schelde-Olesen, M~Kobaek-Larsen,
  A~Koulaouzidis, R~Kroijer, I~Al-Najami, N~Buch, A~Høgh, N~Qvist,
  MK~Thygesen, and U~Deding.
\newblock Choice of colon capsule or colonoscopy versus default colonoscopy in
  {FIT} positive patients in the {Danish} screening programme: a parallel group
  randomised controlled trial.
\newblock \emph{Gut}, 74\penalty0 (10):\penalty0 1616--1623, 2025.
\newblock \doi{10.1136/gutjnl-2024-333687}.

\bibitem[Bromley et~al.(1993)Bromley, Bentz, Bottou, Guyon, Lecun, Moore,
  Sackinger, and Shah]{Bromley1993}
Jane Bromley, James Bentz, Leon Bottou, Isabelle Guyon, Yann Lecun, Cliff
  Moore, Eduard Sackinger, and Rookpak Shah.
\newblock Signature verification using a ``{Siamese}'' time delay neural
  network.
\newblock \emph{International Journal of Pattern Recognition and Artificial
  Intelligence}, 7\penalty0 (4):\penalty0 669--688, 1993.
\newblock \doi{10.1142/S0218001493000339}.

\bibitem[Chopra et~al.(2005)Chopra, Hadsell, and LeCun]{Chopra2005}
S.~Chopra, R.~Hadsell, and Y.~LeCun.
\newblock Learning a similarity metric discriminatively, with application to
  face verification.
\newblock In \emph{2005 IEEE Computer Society Conference on Computer Vision and
  Pattern Recognition (CVPR'05)}, volume~1, pages 539--546, 2005.
\newblock \doi{10.1109/CVPR.2005.202}.

\bibitem[Chen et~al.(2020)Chen, Kornblith, Norouzi, and Hinton]{Chen2020}
Ting Chen, Simon Kornblith, Mohammad Norouzi, and Geoffrey Hinton.
\newblock A simple framework for contrastive learning of visual
  representations.
\newblock In \emph{Proceedings of the 37th International Conference on Machine
  Learning}, volume 119 of \emph{Proceedings of Machine Learning Research},
  pages 1597--1607, 2020.
\newblock URL \url{https://proceedings.mlr.press/v119/chen20j.html}.

\bibitem[Ilina et~al.(2022)Ilina, Ziyadinov, Klenov, and Tereshonok]{Ilina2022}
Olga Ilina, Vadim Ziyadinov, Nikolay Klenov, and Maxim Tereshonok.
\newblock A survey on symmetrical neural network architectures and
  applications.
\newblock \emph{Symmetry}, 14\penalty0 (7):\penalty0 1391, 2022.
\newblock \doi{10.3390/sym14071391}.

\bibitem[Zhang et~al.(2022)Zhang, Qi, Cai, Zhao, Yu, Yue, Yao, and
  Qian]{Zhang2022}
Kai Zhang, Shouliang Qi, Jiumei Cai, Dan Zhao, Tao Yu, Yong Yue, Yudong Yao,
  and Wei Qian.
\newblock Content-based image retrieval with a convolutional {S}iamese neural
  network: Distinguishing lung cancer and tuberculosis in {CT} images.
\newblock \emph{Computers in biology and medicine}, 140:\penalty0 105096, 2022.
\newblock \doi{10.1016/j.compbiomed.2021.105096}.

\bibitem[Schroff et~al.(2015)Schroff, Kalenichenko, and Philbin]{Schroff2015}
Florian Schroff, Dmitry Kalenichenko, and James Philbin.
\newblock {FaceNet}: A unified embedding for face recognition and clustering.
\newblock In \emph{2015 {IEEE} Conference on Computer Vision and Pattern
  Recognition ({CVPR})}, pages 815--823, 2015.
\newblock \doi{10.1109/CVPR.2015.7298682}.

\bibitem[He et~al.(2016)He, Zhang, Ren, and Sun]{He2016_resnet}
Kaiming He, Xiangyu Zhang, Shaoqing Ren, and Jian Sun.
\newblock Deep residual learning for image recognition.
\newblock In \emph{2016 IEEE Conference on Computer Vision and Pattern
  Recognition (CVPR)}, pages 770--778, 2016.
\newblock \doi{10.1109/CVPR.2016.90}.

\bibitem[Hadsell et~al.(2006)Hadsell, Chopra, and LeCun]{Hadsell05}
R.~Hadsell, S.~Chopra, and Y.~LeCun.
\newblock Dimensionality reduction by learning an invariant mapping.
\newblock In \emph{2006 IEEE Computer Society Conference on Computer Vision and
  Pattern Recognition (CVPR'06)}, volume~2, pages 1735--1742, 2006.
\newblock \doi{10.1109/CVPR.2006.100}.

\bibitem[Xiao et~al.(2021)Xiao, Wang, Efros, and Darrell]{Xiao2021}
Tete Xiao, Xiaolong Wang, Alexei~A. Efros, and Trevor Darrell.
\newblock What should not be contrastive in contrastive learning.
\newblock In \emph{International Conference on Learning Representations
  (ICLR)}, 2021.
\newblock URL \url{https://openreview.net/forum?id=CZ8Y3NzuVzO}.

\bibitem[Yang et~al.(2023)Yang, Xiao, Zhang, Guo, Zhao, and Shen]{Yang2023}
Suorong Yang, Weikang Xiao, Mengchen Zhang, Suhan Guo, Jian Zhao, and Furao
  Shen.
\newblock Image data augmentation for deep learning: A survey.
\newblock \emph{arXiv preprint}, 2023.
\newblock \doi{10.48550/arXiv.2204.08610}.

\bibitem[Huang et~al.(2017)Huang, Liu, van~der Maaten, and
  Weinberger]{DenseNet2017}
Gao Huang, Zhuang Liu, Laurens van~der Maaten, and Kilian~Q Weinberger.
\newblock Densely connected convolutional networks.
\newblock In \emph{Proceedings of the IEEE Conference on Computer Vision and
  Pattern Recognition}, 2017.
\newblock \doi{10.1109/CVPR.2017.243}.

\bibitem[Tan and Le(2019)]{EfficientNet2019}
Mingxing Tan and Quoc Le.
\newblock {E}fficient{N}et: Rethinking model scaling for convolutional neural
  networks.
\newblock In Kamalika Chaudhuri and Ruslan Salakhutdinov, editors,
  \emph{Proceedings of the 36th International Conference on Machine Learning},
  volume~97 of \emph{Proceedings of Machine Learning Research}, pages
  6105--6114, 2019.
\newblock URL \url{https://proceedings.mlr.press/v97/tan19a.html}.

\end{thebibliography}

\appendix
\counterwithin{figure}{section}   
\counterwithin{table}{section}    
\titleformat{\section}{\normalfont\Large\bfseries}{Appendix \thesection:}{0.5em}{}

\section{Size and morphology of polyps}
\label{sec:app_A}

\begin{figure}[htbp]
    \centering
    \includegraphics[width=0.8\linewidth]{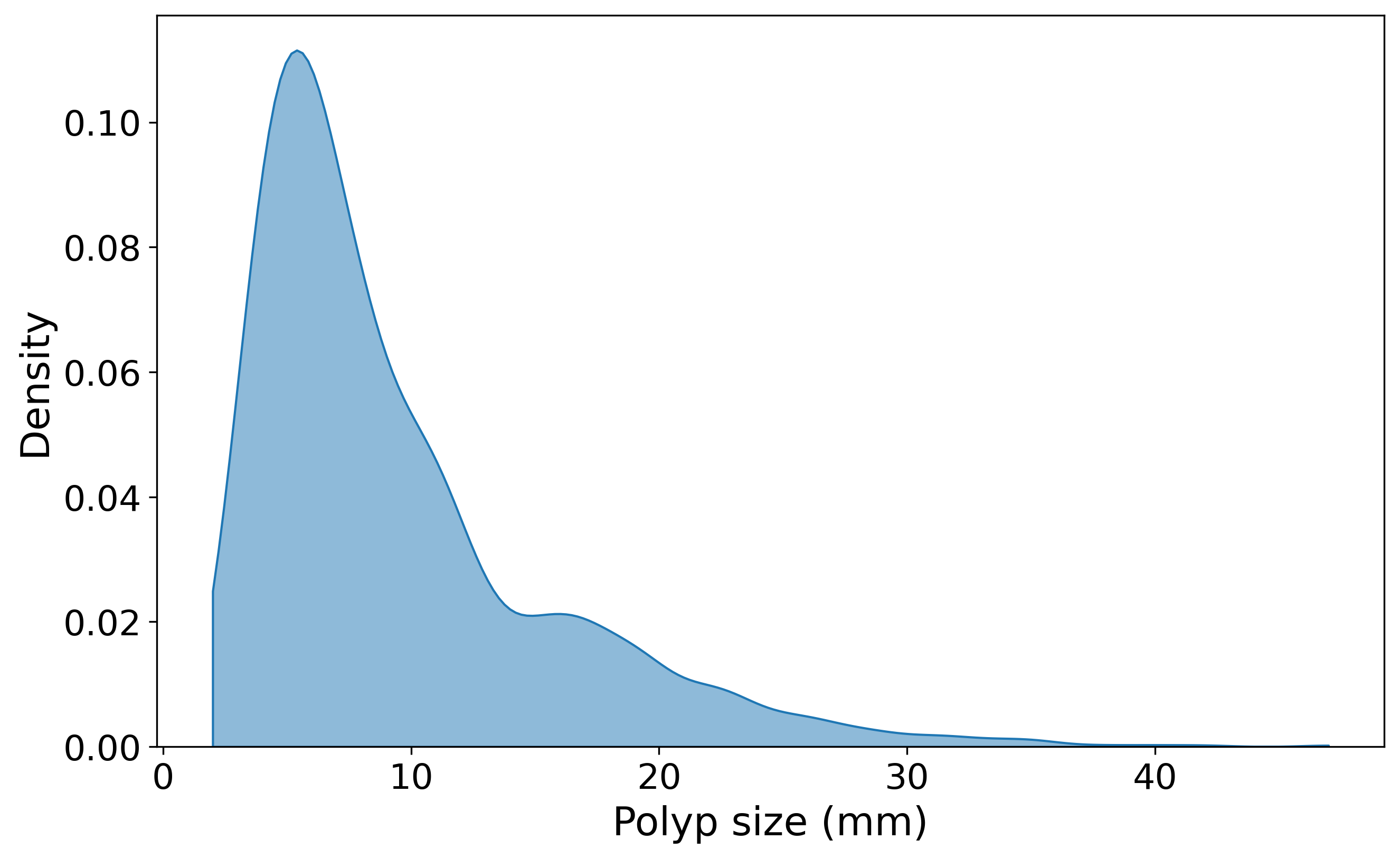}
    \caption{Density plot of polyp size.}
    \label{fig:app_A_1}
\end{figure}

\begin{figure}[htbp]
    \centering
    \includegraphics[width=0.8\linewidth]{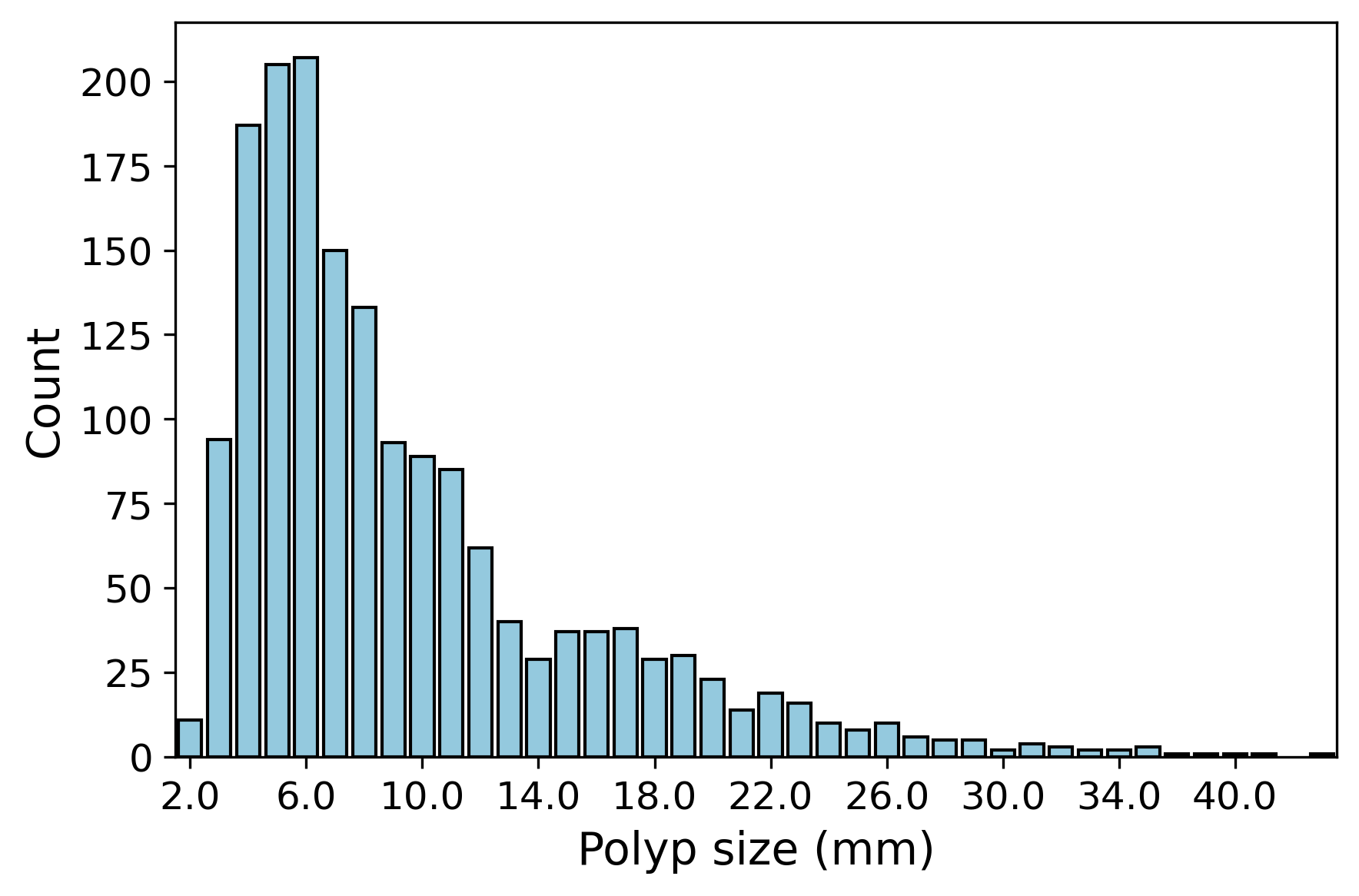}
    \caption{Count plot of different polyp sizes.}
    \label{fig:app_A_2}
\end{figure}

\begin{figure}[htbp]
    \centering
    \includegraphics[width=0.8\linewidth]{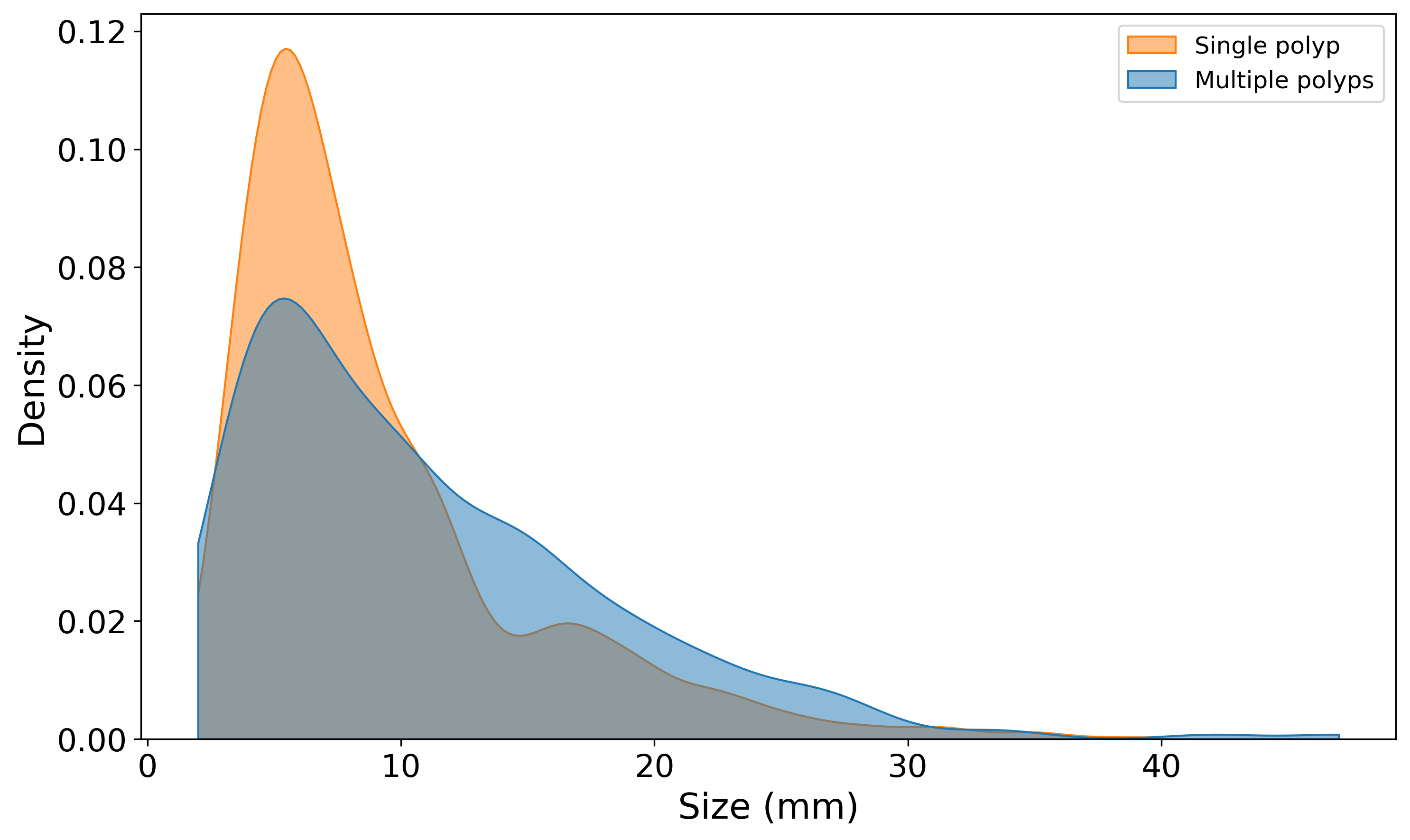}
    \caption{Polyp size for patients with a single polyp versus those with more than one polyp.}
    \label{fig:app_A_3}
\end{figure}

Figures~\ref{fig:app_A_1} and \ref{fig:app_A_2} show density and count plots of the polyp sizes of this study. Figure~\ref{fig:app_A_3} compares the polyp sizes of patients with a single polyp to those with more than one polyp.

A summary of the recorded morphology is displayed in Table~\ref{tab:app_A_1}.
\begin{table}[htbp]
\centering
\caption{Morphology of included polyps.}
\label{tab:app_A_1}
\begin{tabular}{@{} l c @{}}
\toprule
Morphology & Number of polyps \\ 
\midrule
Broad-based/sessile	& 1375 \\
Pedunculated	& 246 \\
Sessile w/mucusol cap & 15 \\
Flat	& 14 \\
Tumor	& 12 \\
Other	& 27 \\
Missing	& 12 \\
\bottomrule
\end{tabular}
\end{table}

\subsection*{Polyp transit times}

Table~\ref{tab:app_A_2} shows the typical transit times (time from first to last sight of a polyp of a single passing). The fractions indicate how many of the polyp transit times are equal to or shorter than the given number of seconds.

\begin{table}[htbp]
\centering
\caption{Polyp transit times.}
\label{tab:app_A_2}
\begin{tabular}{@{} c c @{}}
\toprule
Seconds & Cumulative fraction \\
\midrule
       0     &      0.118 \\
       1     &      0.269 \\
       2     &      0.342 \\
       3     &      0.389 \\
       4     &      0.440 \\
       5     &      0.467 \\
       6     &      0.493 \\
       7     &      0.510 \\
      16     &      0.604 \\
      33     &      0.701 \\
      72     &      0.800 \\
     181     &      0.900 \\
\bottomrule
\end{tabular}
\end{table}


\section{Results using alternative backbones}
\label{sec:app_B}
The hyperparameters that differed from the ResNet setup (Table~\ref{tab:training_params}) were:

\begin{itemize}
    \item For EfficientNet B3 the images were resized from 256x256 to 300x300, the native input size of EfficientNet B3.
    \item EfficientNet B3 and DenseNet121 used learning rate \num{5e-5} while the rest used \num{1e-4}.
\end{itemize}

Tables~\ref{tab:app_B_1} and~\ref{tab:app_B_2} show the cross-validation and final test set performance measures, respectively, corresponding to Tables~\ref{tab:performance_CV} and~\ref{tab:performance_final} for the ResNet backbones.

\begin{table}[htbp]
\centering
\caption{Cross-validation test set performance measures for different backbones. Mean values and $\pm$ 1 standard deviation across the 10 folds. AUC is area under the receiver operating characteristic curve.}
\label{tab:app_B_1}
\begin{tabular}{@{} l ccc @{}}
\toprule
Metric  & DenseNet121 & EfficientNet B0 & EfficientNet B3 \\
\midrule
Accuracy & 0.9172 $\pm$ 0.0092 & 0.9017 $\pm$ 0.0130 & 0.9096 $\pm$ 0.0143 \\
Precision & 0.9218 $\pm$ 0.0144 & 0.8984 $\pm$ 0.0196 & 0.9094 $\pm$ 0.0193 \\
Recall & 0.9121 $\pm$ 0.0116 & 0.9063 $\pm$ 0.0129 & 0.9103 $\pm$ 0.0228 \\
Specificity & 0.9224 $\pm$ 0.0154 & 0.8971 $\pm$ 0.0220 & 0.9088 $\pm$ 0.0222 \\
F1 score & 0.9168 $\pm$ 0.0090 & 0.9022 $\pm$ 0.0123 & 0.9096 $\pm$ 0.0146 \\
AUC & 0.9737 $\pm$ 0.0042 & 0.9673 $\pm$ 0.0094 & 0.9691 $\pm$ 0.0062 \\
Optimal threshold & 0.2310 $\pm$ 0.0207 & 0.2230 $\pm$ 0.0173 & 0.2210 $\pm$ 0.0226 \\
Early stopped epoch & 12.5 $\pm$ 5.1 & 20.0 $\pm$ 8.8 & 12.1 $\pm$ 2.1 \\
\bottomrule
\end{tabular}
\end{table}

\begin{table}[htbp]
\centering
\caption{Test set performance measures for different backbones. AUC is area under the receiver operating characteristic curve.}
\label{tab:app_B_2}
\begin{tabular}{@{} l ccc @{}}
\toprule
Metric & DenseNet121 & EfficientNet B0 & EfficientNet B3 \\
\midrule
Accuracy & 0.9160 & 0.9010 & 0.9128 \\
Precision & 0.9029 & 0.8979 & 0.9238 \\
Recall & 0.9323 & 0.9049 & 0.8997 \\
Specificity & 0.8997 & 0.8971 & 0.9258 \\
F1 score & 0.9174 & 0.9014 & 0.9116 \\
AUC & 0.9745 & 0.9708 & 0.9736 \\
Optimal threshold & 0.20 & 0.20 & 0.22 \\
Early stopped epoch & 9 & 14 & 12 \\
\bottomrule
\end{tabular}
\end{table}

\begin{figure}[htbp]
    \centering
    \includegraphics[width=1\linewidth]{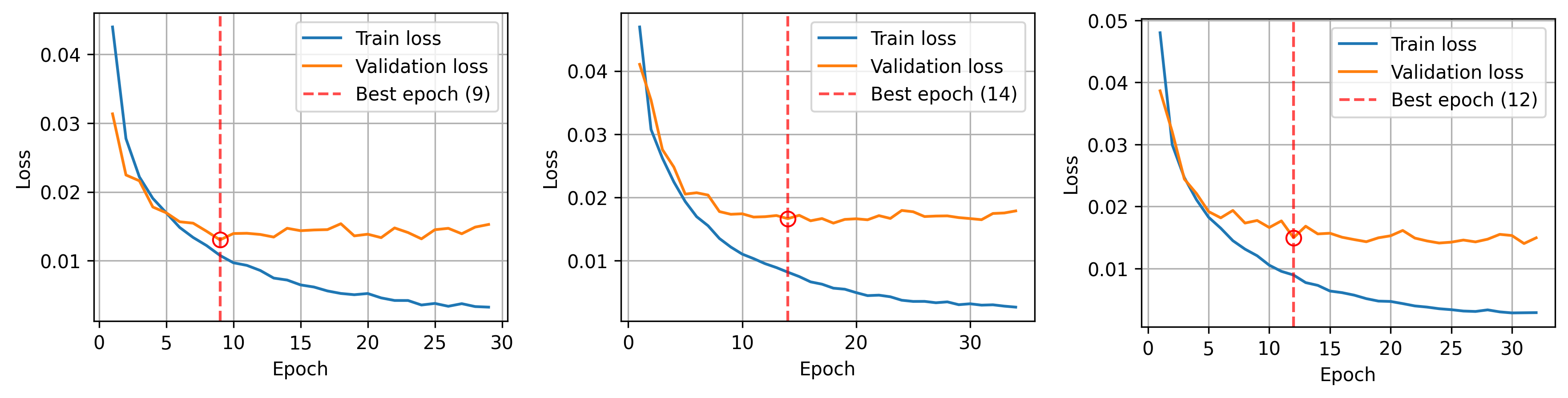}
    \caption{Training and validation loss using cosine distance metrics, and DenseNet121, EfficientNet B0 and EfficientNet B3 from left to right. Best model following triggering of early stopping marked in red.}
    \label{fig:app_B_1}
\end{figure}

\begin{figure}[htbp]
    \centering
    \includegraphics[width=1\linewidth]{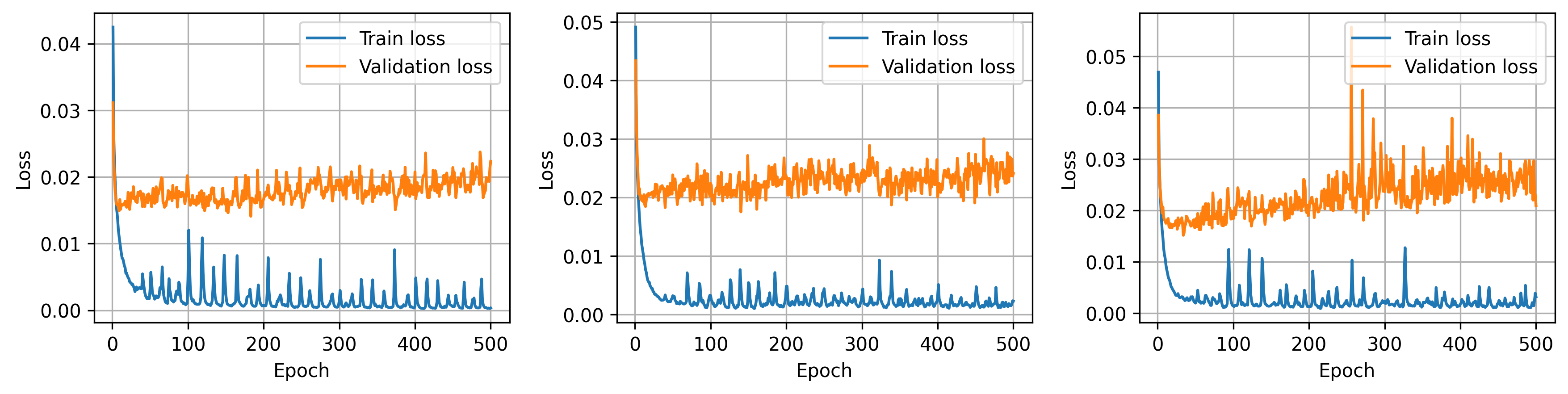}
    \caption{Training and validation loss when  running for 500 epochs. DenseNet121, EfficientNet B0 and EfficientNet B3 from left to right.}
    \label{fig:app_B_2}
\end{figure}

Figures~\ref{fig:app_B_1} and~\ref{fig:app_B_2} show the training curves with early stopping and when running for 500 epochs, respectively, corresponding to Figures~\ref{fig:training_curve_final} and~\ref{fig:training_curve_final_500}.

\section{Distribution of cosine vector distances}
\label{sec:app_C}

\begin{figure}[htbp]
    \centering
    \includegraphics[width=1\linewidth]{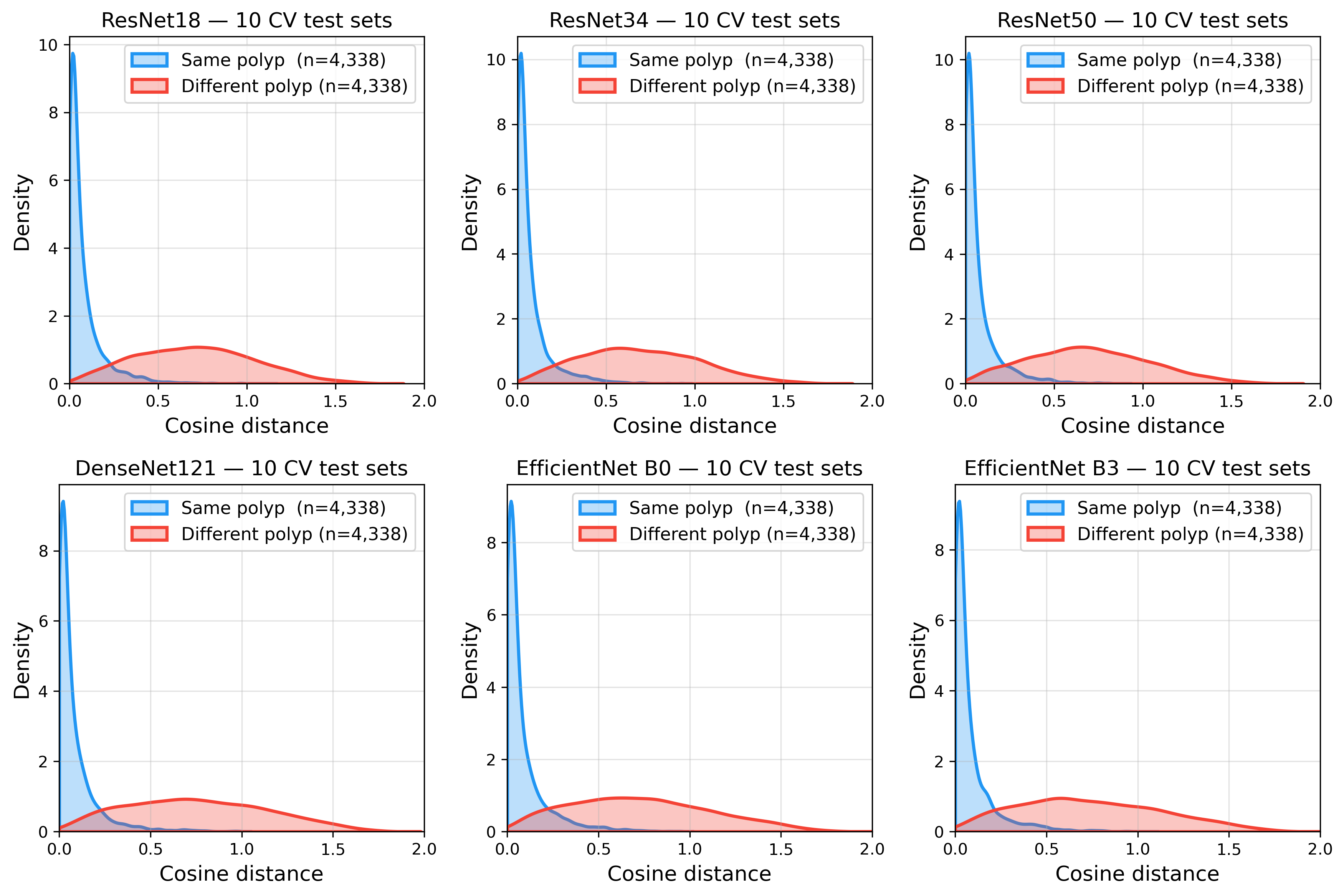}
    \caption{Distribution of cosine vector distances of the 10 test sets of the cross-validation for all six backbones.}
    \label{fig:app_C_1}
\end{figure}

\begin{figure}[htbp]
    \centering
    \includegraphics[width=1\linewidth]{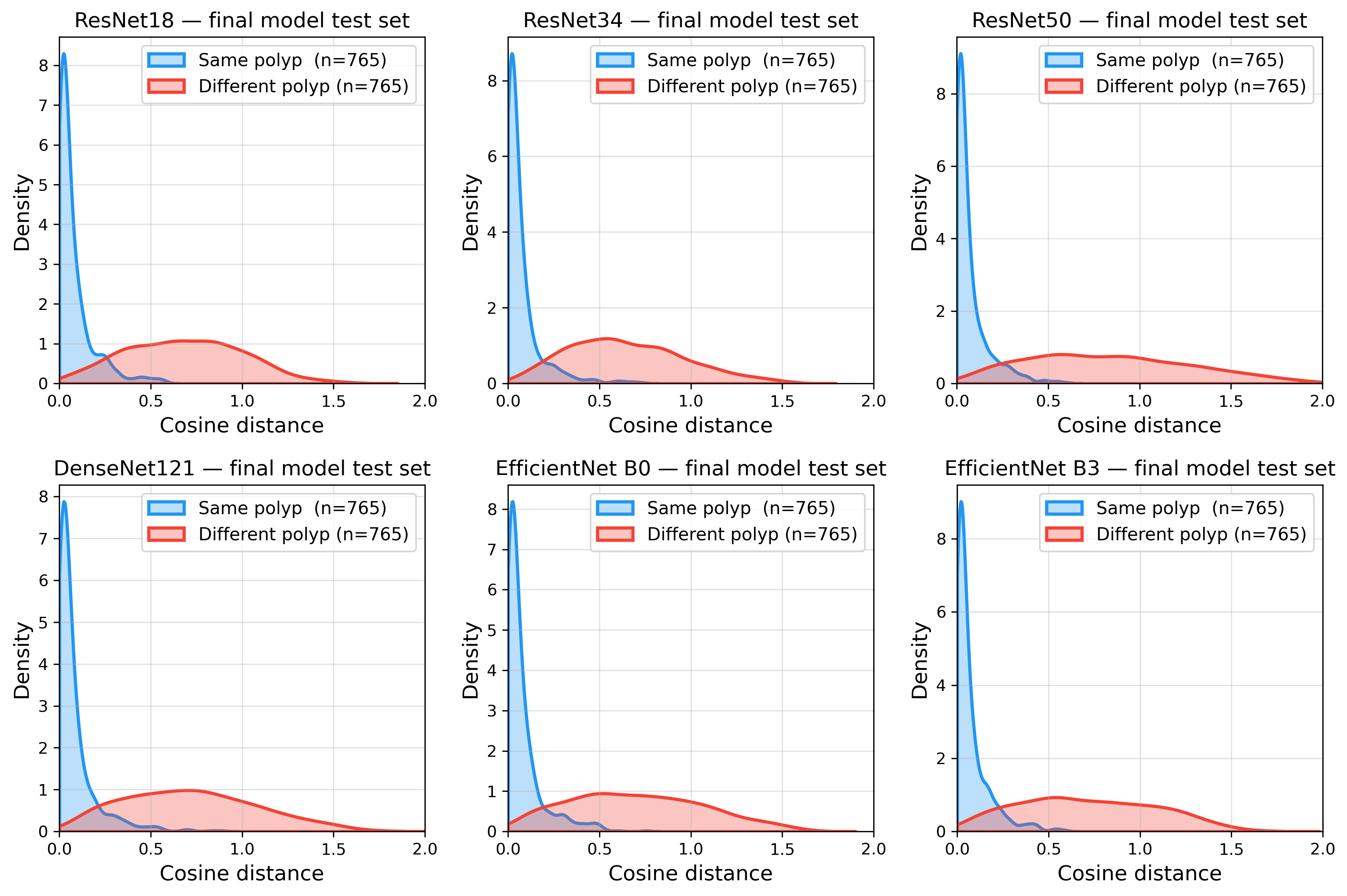}
    \caption{Distribution of cosine vector distances of the final test set for all six backbones.}
    \label{fig:app_C_2}
\end{figure}

Figures~\ref{fig:app_C_1} and~\ref{fig:app_C_2} show the distribution of cosine vector distances for the ``same polyp'' and ``different polyp'' classes for the different backbones for the test sets of the cross-validation setup and for the final full model test set, respectively.

\end{document}